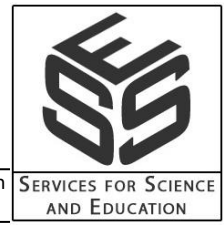

# Redefining Finance: The Influence of Artificial Intelligence (AI) and Machine Learning (ML)

**Animesh Kumar**
Computer Science, Illinois Institute of Technology, Illinois, USA
Computer Technology, Nagpur University, Maharashtra, State, India

## ABSTRACT

With rapid transformation of technologies, the fusion of Artificial Intelligence (AI) and Machine Learning (ML) in finance is disrupting the entire ecosystem and operations which were followed for decades. The current landscape is where decisions are increasingly data-driven by financial institutions with an appetite for automation while mitigating risks. The segments of financial institutions which are getting heavily influenced are retail banking, wealth management, corporate banking & payment ecosystem. The solution ranges from onboarding the customers all the way fraud detection & prevention to enhancing the customer services. Financial Institutes are leap frogging with integration of Artificial Intelligence and Machine Learning in mainstream applications and enhancing operational efficiency through advanced predictive analytics, extending personalized customer experiences, and automation to minimize risk with fraud detection techniques. However, with Adoption of AI & ML, it is imperative that the financial institute also needs to address ethical and regulatory challenges, by putting in place robust governance frameworks and responsible AI practices.

**Keywords:** Artificial Intelligence, Machine Learning, Retail Banking, Predictive Analytics, Fraud Detection, Customer Experience, Ethical Considerations, Regulatory Challenges

## INTRODUCTION

The combination of Artificial Intelligence (AI) and Machine Learning (ML) is one area in which the age-old financial system will witness a revolution - making it more efficient, precise, and innovative. In this paper, we delve into changing the face of finance by AI & ML ranging from personalised customer experience to advanced risk management and much more.

AI and ML have started transforming the financial sector by managing complicated tasks, processing vast data volumes at lightning speed with accuracy, as well as predictive insights that were unattainable earlier with traditional methods. In several ways, these advancements are not only refining operational efficiencies in financial institutions but also improving the decision-making abilities available with more personalized solutions that suit varied needs of clients and stakeholders.

AI and ML are also substantially contributing to effective risk management with intricate algorithms that can quickly detect anomalies, predict market trends by identifying patterns in the data stream, adhering more rigorously than ever to stringent regulatory requirements. To protect themselves from potential threats, the agencies assertively test for vulnerabilities in a financial system based on trust and reliable information.

Moreover, the contribution of AI/ML in finance has upgraded traditional services to tech-enabled solutions with automated investment advisory, trade algorithms and personalized wealth management solutions. These advancements are empoweringfinancial services, enabling both individuals and businesses to make informed decisions and complete their respective fiscal goals more efficiently than ever.

Summing up, it is not merely a technological upgrade but a change in thinking for the finance industry to leverage AI and ML as a transition from traditional mode of doing business which will open unimaginable areas of growth, efficiencies besides customer centricity. As banks and other financial institutions adopt these technologies, they are set to alter the state of finance by opening up new opportunities and creating unprecedented value within an ever more interconnected global economy.

From retail banking to corporate banking, from property and casualty to personal lines, and from portfolio management to trade processing, the next wave of digital disruption in financial services has been unleashed by the concepts and applications of Artificial Intelligence (AI) and Machine Learning (ML). Together, AI and ML are undoubtedly creating one of the largest technological transformations the world has ever witnessed. Within the advanced streams of research in AI and ML, human intelligence blended with the cognitive reasoning of machines is finally out of the labs and into real-time applications.

The Financial Services sector is one of the early adopters of this revolution and arguably much ahead of its leverage compared to other sectors. Built on the conceptual foundations of Innovation diffusion, and a contemporary perspective of enterprise customer life-cycle journey across the AI-value chain defined by McKinsey Global Institute (2017), the current study attempts to highlight the features and use-cases of early-adopters of this transformation. With the theoretical underpinning of technology adoption lifecycle, this paper is an earnest attempt to comment on how AI and ML have been significantly transforming the Financial Services market space from the lens of a domain practitioner, [1-2].

## WHAT IS AI & ML

AI (Artificial Intelligence) and ML (Machine Learning) are the two most powerful words that have revolutionized every industry from finance to healthcare, retail as well more.

**Artificial Intelligence (AI)**
**The Actual Meaning:**
The ability to simulate human intelligence in systems that are capable of thinking and learning like humans. It involves a wide spectrum of strategies and methods processed towards empowering computers to do things which are exclusive to human Intelligence. Few of the areas include optical recognition, speech acknowledgment, senses deduction, decision-making with language understanding.

**AI Can Be Classified into Two Main Streams**
**Narrow AI (Weak AI):**
Category of artificial intelligence that acts optimally when applied to a task constrained within very specific parameters. It includes voice assistants, e.g. Siri or Alexa; recommendation systems and face verification software etc.

**Artificial General Intelligence/AGI (Strong AI):**
Refers to intelligent systems that can understand, learn and apply human intelligence for any task. General AI is actually not a reality but only the concept extension of it.

**Machine Learning (ML)**
ML (Machine Learning) is a type of AI that provides computers with the ability to learn without being explicitly programmed and therefore allow them to predict, make decisions based on data. Unlike traditional programs, which are purposefully built to perform specific tasks, ML algorithms use data (by feeding it back in through the loop) in order to train themselves and make better future decisions.

**Types of Machine Learning**
**Supervised Learning:**
In the case of supervised, an algorithm should learn from labeled data. It is the approach of mapping function from input to output and it learns this mapping by predicting on new data.

**Unsupervised Learning:**
In this, an algorithm is trained on unlabeled data and learns patterns without relying on any form of guidance.

**Reinforcement Learning:**
Reinforcement learning is an area of machine learning that focuses on how actions should be taken in the environment so as to maximize some notion of cumulative reward.

AI and ML are disrupting industries by automating processes, strengthening decision-making ability, and providing a personalized approach to customers. Applications, which range from autonomous vehicles and medical diagnostics to financial trading have transformed the way businesses operate and deliver value in this digital age.

AI and ML have proven to be game-changers for the financial sector. These technologies grant financial institutions the ability to process exabytes of data at unparalleled velocities and levels of accuracy, sustenance by better understanding across multiple dimensions. With AI algorithms, businesses are able to detect anomalies in real-time and optimize investment strategies as well as predict future market trends and intelligent risk management practices. In addition, ML Models also does an outstanding job in everyday tasks e.g., detecting frauds or making customer service easier and thus leaving resources for more strategic plays. This enables organisations to streamline operations and reduce costs, significantly speeding up the time taken for financial services deployment.

Artificial Intelligence (AI) encompasses a broad range of technologies designed to mimic human intelligence, including machine learning (ML), natural language processing (NLP), computer vision, and robotics. Machine Learning, a subset of AI, involves the development of algorithms that enable computers to learn from and make predictions or decisions based on data. ML techniques include supervised learning, unsupervised learning, and reinforcement learning [3]. One of the common and largest implications of AI and ML's influence on business is making interactions with machines smarter and more intelligent (Copeland, 2020; Deloitte Insights, 2019). There has been sizable variance amongst the adoption patterns of the AI and ML-powered transformation across different sectors. The financial services (FS) sector in specific, one of the early adopters and aggressive investors of AI and ML systems (Mazzotta & Chakravorti, 2014) has been realizing the tangible benefits of this transformation. The benefits rangefrom transforming customer experience to creating a more efficient back-office, ensuring regulatory compliance to augmenting the human workforce with timely and relevant actionable insights, and up to helping the enterprises create innovation by enabling a faster go-to-market of new products and solutions (Go et al., 2020), [4-6].

## CURRENT PERSPECTIVE ON AI & ML IN FINANCE

In the hyper competitive digital ecosystem, it appears that Artificial Intelligence (AI) and Machine Learning (ML), have knitted themselves into the very fabric of finance, so essential operations like how decisions are being made and services are delivered to customers.

As the Financial Services sector being more aggressive of the pack, the AI-related investments are, correlating with the 48% cross-sector CAGR, expected to reach $ 10 Bn by 2021 (Deloitte, 2017, 2018b), more interestingly 76% of the Banking CXOs believes that AI adoption will help them create a unique competitive differentiation (Deloitte, 2018b), [7-9].

Here are the few of the areas in Finance where AI & ML are extensively used:
- **Risk Management:** Uses AI and ML algorithms for real-time risk assessment and management. Use cases include analyzing historical data, market trends and external variables to predict risks as well as detect anomalies which allows an optimal risk in the portfolio.
- **Fraud Detection and Prevention:** In financial sectors detecting all those frauds which are happening in monetary transactions with the help of AI (Artificial intelligence) along with ML helps to detect & prevent these fraudulent activities. All such technologies can analyze data, trawling transaction and user behaviors to trace out suspicious activity as it happens, a real-time takedown competence facilitating both financial institutions and their customers.
- **Algorithmic Trading:** The use of AI-driven algorithms is very common in this space where automated systems execute trades based on predetermined metrics and market conditions. The AI powered applications ML models crawl through historical data, news, emotions and market indicators to make rapid trading decisions which alters the approach of conventional buy-and-sell.
- **Credit Rating and Lending:** System powers businesses to predict and rate the likelihood of a customer being able or not able to repay loans using huge datasets including their banking & credit transactions, credit scores, income levels etc to grant loans. These models assist in predicting the exact risk associated with different loans and help financial institutes to provide better lending products from a wide variety of consumers.
- **Investment Management:** AI and ML have revolutionized investment management, AI systems now analyze market trends, asset performance and possible future returns. From personalized investment advice, to managing investments for optimally achieving financial goals and risk levels of the individual investor; all are being handled by efficient ML algorithms.
- **Compliance Management:** Compliance is an extremely critical area and AI technologies help financial institutions stay compliant by automating compliance functions, screening transactions for non-compliance, and providing recommendations to maintain regulations. These AI capabilities help in reducing compliance risks and facilitate regulatory reporting.
- **Quantitative Analysis and Forecasting:** On the basis of historical data, ML models are used to perform quantitative analysis for projecting financial markets as well as recognizing investment opportunities. Based on analysis of large data sets and combined, usually complex relations, these models help predict the future direction in certain markets thus constituting an opinion consumer should be aware of trading strategies or investment advice.
- **Insurance Underwriting and Claims Processing:** An area which is highly time consuming and effort intensive. AI & ML techniques benefit the businesses in several areas including underwriting risk assessment, claims processing automation, fraud claim detection. This advancement helps insurance operations, making it easier and more accurate to determine potential risk of increase while processing claims and the entire process can be faster resulting in a higher customer satisfaction.

To summarise, AI and ML have overhauled the financial sector; providing smarter ways to work (automation), helping in decision making processes with intelligent analysis, improving customer experiences leading to better engagement.

Modern technologies at the forefront of this transformation in the Capital Markets Industry, there is a multitude of opportunities across the industry value chain. Right from Investment and wealth management to the trade processing and back-office and up to the market infrastructure, the technology revolution is expected to be front-lined by AI and ML whereas, although a bit far-sighted, disruption seems inevitable with the evolution of distributed ledger (blockchain) (Lee, 2018). Efficiency, enhanced security, analytical reasoning, and seamless performance will foster the adoption of AI and ML in the Capital Markets Industry (Mendelsohn, 1995), [10] [11].

By incorporating AI and ML-powered reasoning deeply into their business strategies the global capital market majors are already evincing considerable business value of this technology proliferation 1) Financial models have become more data and insight-driven and hence accurate 2) Automated Investment Advisory that is more driven by the data and trends and less of human intuition 3) Automated reporting, integrated workflow, and administration of the market infrastructure 4) faster and responsive deal experience 5) Predictive modeling- Pre and Post-trade, Risk 6) Digital disruption in sales and service. (Accenture, 2018a) [12].AI-empowered smart FinTech has emerged as a sexy and increasingly critical area in AI, data science, economics, finance, and other relevant research disciplines and business domains. This trend was built on the long history of AI in finance, and the new-generation AI, data science and machine learning are fundamentally and seamlessly transforming the vision, missions, objectives, paradigms, theories, approaches, tools and social aspects of economics and finance and driving smart FinTech. AI is empowering more personalized and advanced and better, safer and newer mainstream and alternative economic-financial mechanisms, products, models, services, systems, and applications. This review summarizes the lasting research on AI in finance and focuses on creating a comprehensive, multidimensional and economic-financial problem-driven research landscape of the roles, research directions and opportunities of AI in new-generation FinTech and finance [13].

## REDEFINING THE LANDSCAPE

AI (Artificial Intelligence) and ML (Machine Learning) are changing the face of the Financial sector in more than one way. Few practical implementations include increased prediction accuracy, better risk management and individualized customer service. AI algorithms process large financial datasets in real time, enabling fast pattern and outlier identification missed by humans and enhancing fraud detection and cyber security. At the same time, ML models are playing an increasing role in investment activities and developing stock market predictive models, helping adjust a large corporate portfolio management in line with past performance.

### Personalized Service
AI and ML can be utilized to offer personalized services that are well-suited for an individual interpreter of particular customer requirements, preferences among other things. These offerings can then leverage data analytics to identify consumer usage patterns and financial goals in real time, making personalized recommendations on products or services including investment opportunities over the long-term; both loans and insurance policies.

### Enhanced Accessibility and Convenience
Chatbots and virtual assistants, created with the help of AI solve customer queries at a faster speed and are available 24/7. It can automate responses to FAQs, offer account-based information, and assist in the overall process for a smooth banking experience. Not to miss the endless wait in phone banking which is not eliminated with AI powered customer applications.

### Fast Loan Approval
Where the loan approval takes weeks to conclude due to extensive manual verification, with the help of AI based credit scoring system and machine learning models can provide recommendations within matter of minutes aiding the mandatory manual processing, thereby speedy disbursement of loans.

### Security
AI strengthens fraud detection measures in real-time. Process customer and user behaviors to serve and identify, detect the frauds on their account before it happens by analysing ML algorithms based transactional behaviours.

**Predictive Insights**
Powered by AI, the platform can provide customers with predictive insights of their financial investments and future trends. For example, by anticipating cash flow patterns and advising on saving targets or providing tips related to budget changes in accordance with your spending behavior, artificial intelligence helps you become more knowledgeable regarding finance.

**Better Customer Support**
AI based systems powered by natural language processing (NLP) techniques have become a bridge to enhance customer support. Using voice or text/chat, customers can interact with virtual assistants 24/7 and receive useful answers instantly.

**Customized Investment Advice**
AI powered advisors extend personalized investment advice based on customer risk appetite, financial goals and market conditions. These AI powered platforms recommend diversified portfolios and re-balance investment, automating the kinds of professional investment advice typically only available to the extremely wealthy.

**Trust and Transparency**
AI increases transparency for the consumer by delivering straightforward, understandable explanations of financial products and services. Additionally, ML algorithms can help in regulatory compliance thus protecting the interests of customers with ethical operations for businesses.

**Financial Services Accessibility**
Bringing banking industry and other liquidity services to the underprivileged or unbanked consumers. AI powered applications make the overall process seamless even for customers in remote areas, as these platforms can be extended across various channels such as mobile apps or online portals.

**Product Prediction**
Based on historical and behavioral data, Banks can now offer extremely targeted customer experiences. Using such vast data financial institutions can improve services based on the feedback, anticipate what customers would be most interested in and eventually lead discovery-driven innovation delivering better solutions that meet customer expectations.

Personalized services bridge the gap between a financial institution and its customers and are built on trust. The more we trust the product, the keener we are to disclose our personal information in order to receive a highly personalized service that maximizes consumer value. Artificial Intelligence (AI) can help financial institutions tailor relevant products and services to their customers as well as improve their credit risk management, compliance, and fraud detection capabilities by incorporating chatbots and face recognition systems. The close association between technology and Finance has resulted in the revolution of financial aspects with the help of the implementation of artificial intelligence (AI) (Giudici, 2018). There are three main components of associating AI and Finance with its application: the problems it engenders and the benefits associated. Exploring these components is akin to opening a portal to a world where AI facilitates the comprehension of intricate financial patterns, where challenges are transformed into opportunities for learning, and where favourable outcomes reshape our perceptions of money (ozili,2021), [14-17].

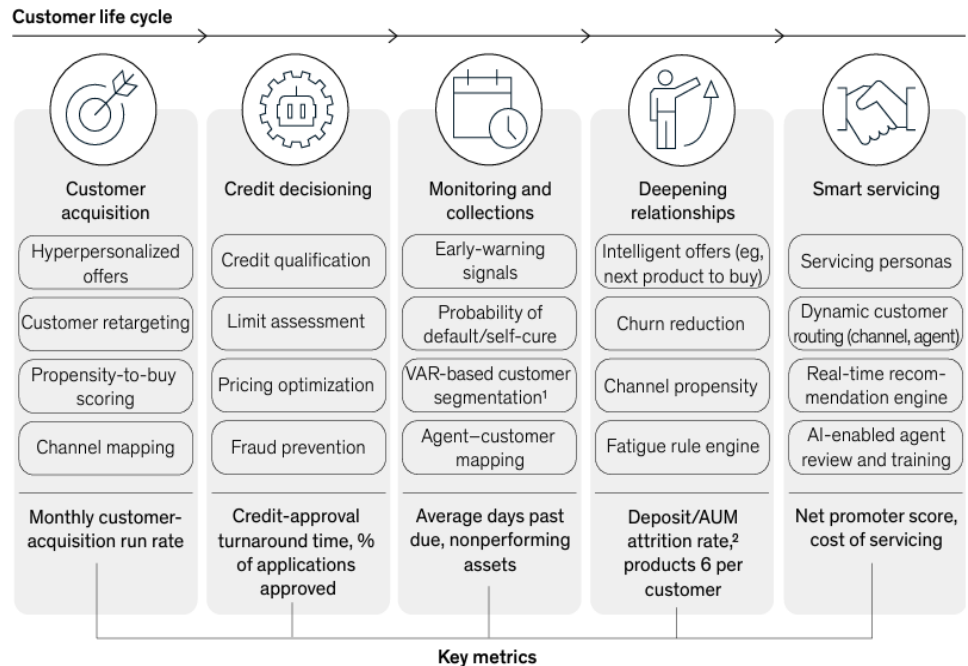

**Figure 1**

Leading banks are relying increasingly on AI and analytics capabilities to add value in five main areas: customer acquisition, credit decisioning, monitoring and collections, deepening relationships, and smart servicing, [18]. The last decade has seen an important rise of data gathering, especially in the financial sectors. Banks are indeed one of the biggest producers of big data, as a matter of fact no other company than the bank has so much data gathered on its customers. Gathering and analyzing this data is a key feature for decision making, particularly in banking sector. One of the most important and frequent decision banks has to make, is loan approval. The challenge is to know how to build a proactive, powerful, responsible and ethical exploitation of personal data, to make loan applicant proposals more relevant and personalized. Machine learning is a promising solution to deal with this problem. Therefore, in the last years, many algorithms based on machine learning have been proposed to solve loan approval issue. However, these algorithms have not taken into consideration Real-time paradigm during processing, [19].

## FUTURE SCOPE

AI (Artificial Intelligence) and ML (Machine Learning) are poised to play a very significant role in both customer experiences as well as operational efficiencies within financial services that will have a massive economic impact on the industry.

Few areas where massive impact could be seen are:
- **Hyper Personalization:** AI and ML will further evolve the personal financial services that are more accurately customized through historical and behavior data. Hyper-personalized data insights leading to predictive financial services, products, and investment strategies may be the next in-line for innovation. This advancement not only increases the rate at which customers satisfaction but also deeper engagement with financial institutions.
- **Security and Fraud Prevention**: As cyber threats continue to grow, AI-based security solutions will improve and can identify anomalies or signs of potential fraud in real-time ensuring steps are taken to prevent them. With new patterns and threats emerging all the time, advanced ML models will continue to lead constantly giving a tough fight with ever-evolving attackers.
- **Automation and Efficiency:** Automation is about to leapfrog into a different paradigm with use of AI & ML powered applications. AI based systems will assist in basic things like account management, customer support or regulatory compliance allowing human resources to focus on more complex, compliance related and strategic roles. This operational efficiency not just cuts the costs but also speeds up administrative processes leading to higher customer engagement.

Emerging Technologies: AI and ML will merge with other emerging technological tools like Blockchain, IoT (Internet of Things), Quantum Computing. Together these integrations will allow for new use cases spanning decentralized finance (DeFi), predictive analytics, asset management and real-time transaction surveillance. The Next Generation of Financial Services Industry will be reimagined by joint initiatives from fintech startups, established financial institutions & tech giants.

The layers of the AI-bank capability stack are interdependent and must work in unison to deliver value, as discussed in the first article in our series on the AI bank of the future. In our second article, we examined how AI-first banks are reimagining customer engagement to provide superior experiences across diverse bank platforms and partner ecosystems Hyper-personalization process with learning (ML) and artificial intelligence (AI) techniques for marketing functions like segmentation, targeting, and positioning based on real-time analytics throughout the customer journey and key factors driving effective customer-centric marketing. with varied customer demands and preferences, it's imperative for businesses to cater to their needs and sustain growth. Use of AI enables hyper-personalized customer experience which is extended by analyzing vast amounts of data to understand customer's individual preferences & map their behaviour [20-23]. Manual fraud detection has become inefficient due to the intricate nature and high volume of transactions. Artificial intelligence (AI) is already a very efficient technology for detecting and preventing financial crime [24]. Generative AI can be used in the financial sector for generating financial models, predicting market trends, and optimizing investment strategies. It can also assist in fraud detection and risk assessment, [25,26].

## CONCLUSION
To conclude, the future of AI (Artificial Intelligence) and ML (Machine Learning) in finance has immense potential to revolutionize the finance industry. These advanced technologies have major implications for how financial institutions engage with customers, operate and innovate in an increasingly digital world.

Personalized customer experiences will be lifted to new heights with AI and ML, delivering bespoke financial advice, predictive insights as well as seamless interaction thanks not only to advanced automation but with AI-driven conversation modules. This increased degree of customization extends not only to higher satisfaction but also strengthens engagement and loyalty.

Furthermore, AI will only improve the security measures that already were implemented in a financial institution with real-time fraud detection and anomaly recognition as well as deeper protection strategies. While cyber threats are getting more and more sophisticated, the capability of AI to plough through mountains of data looking for patterns will be essential in ensuring that customer assets remain protected.

In the end, with a great pace of evolution and penetration into different layers in financial services, Machine Learning & Artificial Intelligence will act as an enabler providing efficiency and innovation. Doing so responsibly will also be more important in order to fully realize the potential for value creation, social good and sustainable growth of these technologies within the international financial ecosystem.